\documentclass{article}

\usepackage{cite}
\usepackage{amsmath,amssymb,amsfonts}
\usepackage{algorithmic}
\usepackage{graphicx}
\usepackage{soul}
\usepackage{textcomp}
\usepackage{subcaption}
\usepackage{amsmath}

\usepackage{tikz}
\usetikzlibrary{shapes,arrows}
\usetikzlibrary{decorations.markings}
\usetikzlibrary{calc}
\usetikzlibrary{arrows.meta}
\usetikzlibrary{positioning}

\tikzstyle{block} = [draw, rectangle, inner sep = 0cm, minimum height=1.0cm, minimum width=1.0cm, very thick]
\tikzstyle{summer} = [draw, circle, inner sep = 0.0cm, minimum width=3mm, very thick, draw] 

\newcommand{\wn}{\omega_n}

\title{Dynamical System Parameter Identification\\using Deep Recurrent Cell Networks}

\author{Erdem Akagündüz \and Oguzhan Cifdaloz}

\begin{document}
\maketitle

\begin{abstract}
In this paper, we investigate the parameter identification problem in dynamical systems through a deep learning approach. Focusing mainly on second-order, linear time-invariant dynamical systems, the topic of damping factor identification is studied. By utilizing a six-layer deep neural network with different recurrent cells, namely GRUs, LSTMs or BiLSTMs; and by feeding input/output sequence pairs captured from a dynamical system simulator, we search for an effective deep recurrent architecture in order to resolve damping factor identification problem. Our study’s results show that, although previously not utilized for this task in the literature, bidirectional gated recurrent cells (BiLSTMs) provide better parameter identification results when compared to unidirectional gated recurrent memory cells such as GRUs and LSTM. Thus, indicating that an input/output sequence pair of finite length, collected from a dynamical system and when observed anachronistically, may carry information in both time directions for prediction of a dynamical systems parameter.
\end{abstract}


\section{Introduction}
\label{sec:introduction}
System identification describes a set of methods, which uses experimental input/output data from a system, in order to identify its dynamical properties. Depending on the class of systems under inspection, there are a number of approaches that may be applied to system identification \cite{Bekey.1970,Astroem.1971,Ljung.1999}. While non-parametric methods  \cite{Rake.1980,Wellstead.1981,Marple.1987,Ljung.1999} try to estimate a generic model from step responses, impulse responses, frequency responses, etc., parametric methods \cite{Astroem.1965,Box.1970,Natke.1982,Helmicki.1991,Gu.1992,Ljung.1999} aim at estimating parameters within a user-specified model. 

The steps taken to identify a system may be generalized as follows: obtaining experimental data; determining a structure for the model; devising a criterion for model fitting; estimating the parameters; and finally, model validation. System identification methods utilize different mathematical models in order to achieve its objectives. Models can be continuous-time (differential) equations, discrete-time (difference) equations, or a hybrid combination. Models can also be described in a variety of ways. Use of state-space models based on transfer functions are common. Generating experimental data involves several approaches. However, input signals that excite  all the relevant frequencies of a system is an important factor. Identification can be achieved either on-the-fly or offline. The identification of parameters and the use of identified parameters on-the-fly in order to update controller parameters is highly related to adaptive control schemes.

In this paper we tackle the problem using a machine learning approach. By utilizing different deep recurrent neural network (DRNN) architectures and defining the issue as a sequence regression problem, we aim at finding the most practically effective architecture. We compare different gated recurrent cells and then analyze at what exact instant and with what kind of an input the system should be excited in order to obtain the best parameter identification results.

The remainder of this paper is organized as follows: the following subsections introduce the problem statement and the related literature. Section 2 describes the dynamical systems model for which the parameters are to be identified. Section 3 presents details of the deep recurrent neural network model used to solve the problem. Section 4 details the experimental setup, including the simulation environment, the extent of the dataset created, and the preprocessing techniques used for training the model’s input. Section 5 presents a discussion of the experimental results. Finally, Section 6 concludes the paper and outlines potential future research.

\subsection{Problem statement}
This paper is focused on parameter identification of second order, linear time-invariant dynamical systems. These systems can be described as second order transfer functions. A general second-order system has two poles, which can be both real or complex conjugate. Complex conjugate poles are associated with two parameters: natural frequency and damping factor. In this paper, it is assumed that the poles are complex conjugates, and the study aims at identifying the damping factor. 

Damping factor is largely an uncertain parameter, especially for electromechanical systems with small inertia/spring and inertia/friction ratios \cite{Hilkert.2011,Miller.2013}. In such systems, although inertia is a relatively known factor, considerable uncertainty is associated with the spring and friction constants. In this paper, the objective is to identify the damping factor through the use of machine learning algorithms. One important aspect of the method in question is that the algorithms are unaware of the model structure. The only information fed to the algorithms is that a certain input/output pair is associated with a fixed damping factor.

\subsection{Related Literature}
There has been growing interest in the combination of machine learning techniques, specifically recurrent neural networks and dynamic system modeling/identification since the first introduction of recurrent neural networks in  \cite{Pearlmutter89,Cleeremans89,Richard.2019}. For a detailed overview on the subject, see  \cite{richard2019,Chiuso19}. The earlier approaches used standard RNNs to attack the problem, which yielded some promising but limited results \cite{NarendraParthasarathy.1990,PhamLiu.1995,Mohajerin96,Wang06,RubioYu.2007,DinhBhasinDixon.2010}. The limiting factor with the RNNs, not only in dynamical systems modeling but also in general, was their difficulty in training to learn short- or long-term dependencies within the input sequence. Some other hybrid RNN architectures such as artificial deep belief networks were also utilized  \cite{Cheon.2015} for the problem, but they also suffered similar limitations. The main reason for this limitation was that backpropagation deep in time (i.e., sequence dimension) gave rise to the so-called ``\emph{vanishing gradients}'' problem, which was first mentioned in 1991 by \cite{Hochreiter91}. In order to overcome this, the same author proposed a specific RNN architecture with gated units in 1997, named Long-Short-Term Memory (LSTM) \cite{Hochreiter1997LongSM}.

The idea of the LSTM was to create a memory cell within the RNN architecture by ensuring a constant error overflow, so that long- (or short-) term dependencies were not lost during backpropagation in time. LSTM attracted significant praise and was applied to numerous sequence modeling problems, including dynamical system modeling and identification \cite{Wang.2017}. Different versions of recurrent cell networks were also proposed such as convex-based LSTM \cite{Wang.2017}, the gated recurrent unit (GRU) \cite{Cho_2014} and also bidirectional (bi-)LSTM \cite{GRAVES2005602}.

Alternatively, deep convolutional neural networks (CNN) have also been utilized in the literature \cite{Genc.2017,Andersson.2019} for the same problem. Since CNN architectures lack the ability to create time dependencies (i.e., memory), these approaches are uncommon when compared to RNN-based techniques. However, we believe that unexplored feature extraction capabilities of CNNs from dynamical system sequence data are still likely to attract more attention.

Training deep recurrent neural networks that can be successfully utilized for any type of intelligent decision including system identification, is an active field of research \cite{_Hermans2013,_Pascanu2014}. Last few years have shown an increasing trend on studies that focus on applying deep learning techniques to systems identification problem \cite{_9003133,_Ayyad2020,_Brusaferri2019,_Kumar2020,_Lin2014,_Mastorocostas2002,_Passalis2020,_Tavoosi2013}. Some of the recent methods even utilize LSTMs for the problem \cite{_Gonzalez2018}. However, the literature contains no approved or generally accepted off-the-shelf deep learning architecture for dynamical systems parameter identification. Although there have been comparative studies published such as \cite{ogunmolu2016}, which utilized a multi-layered artificial neural network (ML-ANN), an RNN, a single LSTM cell, and a single GRU cell to assess performance in nonlinear systems identification, there remains unanswered questions such as ``How deep a network is needed?'', ``Which types of layers are needed?'' and ``What kind of a recurrent cell performs better?'' for the dynamical systems identification problem.

\section{Dynamic Model Description}
\begin{figure}[t]
\begin{subfigure}{5.6cm}
    \includegraphics*[trim= 0 0 0 0 ,clip=true,width=1\textwidth]{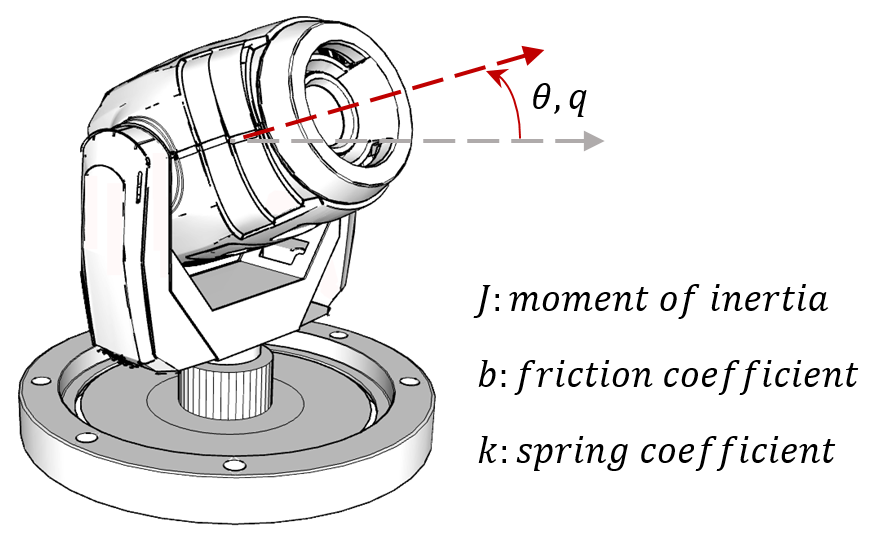}
\end{subfigure}	
\begin{subfigure}{2.86cm}	
\begin{tikzpicture}[auto, >=latex', node distance = 1.5cm]
    \node[inner sep=0pt] at (0,0) (u) {}; 
	\node[summer, name=summer1, right of=u, node distance = 1.0cm] {};
    \node[block,  name=J,       right of=summer1, node distance = 1.5cm] {\Large $1/J$};
    \node[block,  name=int1,    right of=J, node distance = 2.0cm] {\Large $\int$};
    \node[inner sep=0pt, right of=int1, node distance = 1.5cm] (node1) {}; 
    \node[block,  name=b,       above of=J] {\Large $b$};
    \node[summer, name=summer3, above of=summer1] {};
    \node[block,  name=k,       above of=b] {\Large $k$};
    \node[block,  name=int2,    right of=k, node distance = 2.0cm] {\Large $\int$};
    \node[inner sep=0pt, right of=int2, node distance = 1.5cm] (node2) {}; 
    \node[fill=red, inner sep=0pt] at ($(node1)!.5!(node2)$) (node3) {}; 

    \draw[->, >={Stealth[length=2mm]}, very thick] (u) -- node {$u$} (summer1);
    \draw[->, >={Stealth[length=2mm]}, very thick] (summer1) -- (J);
    \draw[->, >={Stealth[length=2mm]}, very thick] (J) -- node {$\dot{q}$} (int1);
    \draw[->, >={Stealth[length=2mm]}, very thick] (int1) -- node {$q$} (node1) -| (node2) |- (int2);
    \draw[->, >={Stealth[length=2mm]}, very thick] (int2) -- node {$\theta$} (k);
    \draw[->, >={Stealth[length=2mm]}, very thick] (node3) -- (b);
    \draw[->, >={Stealth[length=2mm]}, very thick] (k) -| (summer3);
    \draw[->, >={Stealth[length=2mm]}, very thick] (b) -- (summer3);
    \draw[->, >={Stealth[length=2mm]}, very thick] (summer3) --  node[pos=0.95, yshift=0mm] {$-$}(summer1);
\end{tikzpicture}
\end{subfigure}	
\caption{An illustration of a gimbal (left) and its linear model block diagram (right).}
\label{fig:gimbal_bd}
\end{figure}

The dynamic model considered in this paper is described by a second order, linear differential equation. Second order linear differential equations arise in a variety of systems, such as in rotational dynamics, which are described in their general form as
\begin{align}
    J\ddot{\theta} + b\dot{\theta} + k\theta = u
\end{align}
where $J>0$ denotes the moment of inertia, $b>0$ denotes the viscous damping (friction) coefficient, $k>0$ denotes the spring constant, $u$ denotes the externally applied torque, and $\theta$ denotes the angle of rotation. In this paper, linearized rotational dynamics of a gimbal are considered. A simple depiction and the model block diagram representing the system is illustrated in Figure~\ref{fig:gimbal_bd}.

The differential equation that describes this system from the torque input $u$, to the angle output $\theta$ is given in state-space form by
  \begin{align}
    \left[\begin{array}{c}
        \dot{q} \\[2mm]
        \dot{\theta}
    \end{array}\right] =
    \left[\begin{matrix}
        -b/J & -k/J \\[2mm]
          1  &   0
    \end{matrix}\right]
    \left[\begin{array}{c}
        q \\[2mm]
        \theta
    \end{array}\right] +
    \left[\begin{matrix}
         1/J\\[2mm]
          0 
    \end{matrix}\right]
    u, \qquad
    \theta =
    \left[\begin{matrix}
         0 & 1
    \end{matrix}\right]
    \left[\begin{array}{c}
        q \\[2mm]
        \theta
    \end{array}\right]
 \end{align}
where $q \triangleq \dot{\theta}$ represents the angular rate.
The transfer function from the input $u$, to the output $\theta$ is given by
 \begin{align}
   T_{u \theta}= \frac{1/J}{s^2 + b/J s + k/J}
   \label{equ:T_u_theta}
 \end{align}

This transfer function can be represented as a standard second order system multiplied by a constant gain, $\alpha$, and given by:   
 \begin{align}
   \frac{\alpha\omega_n^2}{s^2 + 2 \zeta \omega_n s + \omega_n^2}
   \label{equ:second_order_system_tf}
 \end{align}
where $\omega_n^2 \triangleq k/J$, $2 \zeta \omega_n \triangleq b/J$, and $\alpha \triangleq 1/k$. In a standard second order system, $\omega_n$ denotes the natural frequency and $\zeta$ is called the damping factor.

Second order systems may have complex conjugate poles, and hence their impulse responses may be oscillatory. The poles of a standard second order system given in Equation~\ref{equ:second_order_system_tf} are at
\begin{align}
    s_{1,2} = -\zeta \omega_n \pm j \omega_n \sqrt{1-\zeta^2}
\end{align}
In terms of the system parameters, the natural frequency and the damping factor are given by
\begin{align}
  \omega_n = \sqrt{\frac{k}{J}} 
  \quad \mathrm{and} \quad
  \zeta = \frac{b}{\sqrt{4kJ}}.
  \label{equ:omegan_zeta}
\end{align}
Since all the parameters ($J,b,k$) of the system are positive, from Equation~\ref{equ:omegan_zeta}, it can be seen that $\zeta$ can not take negative values or zero. $\zeta = 0$ implies $b = 0$, i.e. a system with zero friction. However small, friction always exists in a mechanical system such as described in this paper. {$\zeta = 0$ may also imply $k \rightarrow \infty$ or $J \rightarrow \infty$. Both of these two conditions physically correspond to a rigid system, with no freedom of rotation.}\\

Hence, for the system in consideration we may safely assume that $\zeta > 0$. With this assumption, two cases are of importance:
\begin{enumerate}
    \item For $\zeta \geq 1$ (i.e., $b^2 \geq 4kJ$), the transfer function given in Equation~\ref{equ:second_order_system_tf} can be described as two first order systems connected in series. Both poles are stable real poles. This case is excluded from the study because in such a series combination, the damping factor loses its oscillatory effect on the system and there will be no overshoot.
    \item For $0 < \zeta < 1$ (i.e. $b^2 < 4kJ$), the poles of the transfer function given in Equation~\ref{equ:second_order_system_tf} are stable complex conjugate poles. In this study, it is aimed to identify the damping factor which results in (undesired) oscillations. This case is addressed in this paper.
    Naturally, when $0.5<\zeta<1$, oscillations die out relatively quickly. \emph{However, one of the reasons why $0<\zeta<1$ is targeted in this study is that there may be a large and varying uncertainty associated with $\zeta$ that the control system designers would like to know in order to increase the performance of their designs and not have to design a conservative controller.}
\end{enumerate}

The unit step response of the second order system given in Equation~\ref{equ:second_order_system_tf} is given by
\begin{align}
	y(t) =  \alpha\left[1 - \frac{e^{-\zeta\wn t}}{\sqrt{1-\zeta^2}} \sin{\left( \omega_d t + \tan^{-1}{\frac{\sqrt{1-\zeta^2}}{\zeta}}\right)}\right] 
\end{align}
where $\omega_d \triangleq \wn\sqrt{1-\zeta^2}$.

It is noted that the damping factor, $\zeta$, impacts all aspects of the step response, namely, the time constant, frequency, phase, and overshoot. Specifically, because of its impact on oscillations in the system response, estimating $\zeta$ as accurately as possible is critical. However, in a real physical system, estimating $\zeta$ can prove to be difficult, due to uncertainties in the parameters and variations of parameter values in time.  Temperature variations are notorious for their effect on parameter values. One other important reason for parameter uncertainty is the mass manufacturing process. In practice, parameters are often identified for a handful of prototypes and a control design is based on those identified parameters, and implemented on all manufactured devices. This process results in undesired variations in closed loop performance characteristics among products. The procedure described in this paper proposes a method to identify system parameters for an automated custom control design for each device.

\section{Deep Learning Model}
The model proposed in this study that predicts a parameter of a given dynamical system is a deep recurrent neural network (DRNN). DRNNs are extensions of RNNS with additional layers such as non-linearity layers (e.g., ReLU layers), decision layers (e.g., {dense, fully-connected layers}), regularization layers (e.g., drop-out layers), etc. The model we propose in this paper is a 6-layered DRNN  (see Table \ref{DeepLModel}) that predicts a dynamical system parameter (damping coefficient $\zeta$ in our case) of the model, when a $\Delta$ second ($\Delta$ being 3 seconds in our experiments) input/output (I/O) sequence pair is fed into it. 

As can be seen in Table~\ref{DeepLModel}, the first layer of the proposed DRNN architecture is an input layer that accepts an input/output sequence pair of $\Delta$-seconds. The size of the input/output sequence pair that is fed into the model is 2$\times$($\Delta\cdot f_{s}$), with $f_{s}$ being the sampling frequency of the system and $\Delta\cdot f_{s}$ being the sequence length. However, as can be seen in Table~\ref{DeepLModel}, the input fed into the recurrent cell of the DRNN model is sizes 168 $\times$ 11, with 11 being the sequence length and 168 (42$\times$4) being the feature vector size. As explained in detail in the next subsection, at the initial layer of the proposed DRNN architecture, a frequency domain transform is applied to the $\Delta$-seconds input/output sequence pairs, in order to obtain a 168 $\times$ 11 sequence. This frequency domain sequence is the actual input fed into the recurrent cells of the succeeding layer.

The ability to automatically construct a set of feature extractors, in a data-driven and task-dependent manner, is the trademark of deep neural networks (DNNs). Accordingly, one may argue that instead of utilizing an initial handcrafted transform layer, as in the proposed architecture, a set of convolutional layers can provide the necessary transforms (i.e., features) needed for the system. However, we chose to use a frequency transform as the first layer, mainly due to two reasons. First, the parameter that we aim at predicting is a coefficient of the Laplacian transform of a dynamic system, which we assume to be linear. Thus, transforming the input into a frequency domain, heuristically leads to creating a supposedly linearly separable feature space. Second, by using a frequency transform, we shorten the sequence length, which in turn reduces the effective depth of the architecture. The depth of a deep neural model is one of main factors that helps in its training for a desired task \cite{Szegedy2015GoingDW,He_2015,Glorot2010UnderstandingTD}. Although it is the depth of a DNN that helps it create abstract features from the data, depth is also responsible for certain problematic training issues such as vanishing gradients \cite{Hochreiter91}. While we refer to the proposed architecture as ``6-layered,'' the depth of an RNN can be considered ambiguous. During training, the RNN layer is propagated in ``time,'' hence the depth is effectively related to the sequence length (or the ``truncation length'' if the backpropagation in time is truncated \cite{Werbos90,Pascanu13,SutskeverPHD13}). If we use a shorter frequency domain sequence (such as size 11), instead of a $\Delta\cdot f_{s}$ time sequence (which is practically 3,000 in our case), the actual depth of the backpropagation is significantly reduced and training the DRNN becomes a much easier task. Regarding the details of the input layer frequency transform, please refer to the next subsection. 

\begin{table}[t]
\centering
\begin{tabular}{|l|l|l|}
\hline
\textbf{Layer No.}  & \textbf{Type} & \textbf{Layer Properties}\\ \hline
1 & Input Transform & \begin{tabular}[c]{@{}l@{}}Frequency Domain Transform \\\emph{(42$\times$4) $\times$ 11 tensors}\end{tabular}\\ \hline
2 & RNN Cell        & \begin{tabular}[c]{@{}l@{}}RNN Cell\\ (GRU, LSTM, or BiLSTM) \\\emph{with 256 hidden units}\end{tabular} \\ \hline
3 & Fully Connected & \begin{tabular}[c]{@{}l@{}}Fully Connected Layer\\\emph{256 nodes}\end{tabular}\\ \hline
4 & ReLU            & ReLU non-linearity Layer\\ \hline
5 & Dropout         & \begin{tabular}[c]{@{}l@{}}Dropout Layer\\\emph{50\% drop}\end{tabular}\\ \hline
6 & Fully Connected & \begin{tabular}[c]{@{}l@{}}Fully Connected Layer\\\emph{single output node}\end{tabular}\\ \hline
- & Regression      & Mean-squared-error loss\\ \hline
\end{tabular}
\caption{The proposed deep recurrent neural network model.}
\label{DeepLModel}
\end{table}

The input layer is followed by the recurrent layer of the network. The recurrent layer can be any type of a recurrent cell that accepts sequence inputs such as a GRU \cite{Cho_2014}, an LSTM \cite{Hochreiter1997LongSM}, or a BiLSTM \cite{GRAVES2005602} cell. In our experiments, these three types of recurrent cells were bench-marked in order to discover which recurrent cell was the more successful in predicting dynamical system parameters. Details on these recurrent cells are provided in Section \ref{rcell}.

Any recurrent cell can be designed to output a single value. Hence, the first two layers are technically sufficient to create a parameter prediction (i.e., regression) network. However, in our DRNN model, the recurrent layer is succeeded by a fully-connected layer. The hidden units (totaling 256 nodes) of the recurrent cell are fully connected to this dense layer so that more complex features can be obtained using the hidden units (also referred to as the ``states'') of the recurrent cell. Additionally, this fully-connected layer is succeeded by a rectified linear unit (ReLU) \cite{NairRELU10}, with the purpose of creating non-linearity within the decision space. Moreover, the ReLU layer is followed by a drop-out layer \cite{Srivastava14} for the purposes of regularization, and for avoiding the issue of feature overfitting. Finally, another fully-connected layer, this time having a single output value which provides the parameter value to be predicted, was appended to the network. During training, an L2 norm regression layer was used to feed the back-propagated derivatives to the stochastic gradient descent optimizer.

In summary, the proposed DRNN architecture is a dynamical system para-meter prediction network, with the fundamental properties being transforming a time signal to the frequency domain, utilizing recurrent operation within a sequence, convolving advanced features, providing nonlinearity in the feature space and avoiding overfitting. The training details of the proposed DRNN are provided in Section \ref{ExpSec}.

\subsection{Input Sequence}
\label{InpSeq}
The recorded input/output pair sequences are time signals. However, the sequence that we fed into the proposed DRNN model was a frequency domain representation of this time signal, which was obtained by utilizing the Short-time Fourier Transform (STFT). STFT is a sequence of Fourier transforms of a windowed signal. Instead of providing the frequency information averaged over the entire signal time interval (like the standard Fourier transform), STFT provides the time-localized frequency information for situations, in which frequency components of a signal vary over time. STFT are widely used as features for time signals in many intelligent signal processing applications such as audio signal processing \cite{Ghoraani2011}. 

\begin{equation}\label{stft}
    X_{STFT}[m,n]=\sum_{k=0}^{L-1}{x[k]\cdot g[k-m]\cdot e^{-j2\pi nk}}
\end{equation}   

 In Equation (\ref{stft}), $x[k]$ denotes a signal and $g[k]$ denotes an L-point window function, hence STFT of $x[k]$ can be interpreted as the Fourier transform of the product, $x[k]\cdot g[k–m]$. Consequently, the calculated $_{STFT}[m,n]$ is a 2 dimensional complex matrix, where the first dimension represents time, and the second dimension represents sample frequencies. The finite size of the window function $g[k]$ and how much overlap each neighboring frame has, designates the sequence size of the calculated $X_{STFT}[m,n]$.  
 
 In our model, $x[k]$ is a $\Delta\cdot f_{s}$ long sequence with 2 dimensions, that belong to the input/output sequence pair, for both of which STFT are calculated separately. The size of the window function is selected as $L = k\cdot\Delta\cdot f_{s}$, with $p$ inter-frame overlap, therefore creating a $\frac{1-k}{k(1-p)}+1$ long sequence in the STFT time dimension. In the frequency dimension, the range is limited to a certain interval with exponential sampling of N distinct frequency values; ergo, creating two complex sequences of size $N \times (\frac{1-k}{k(1-p)}+1)$ (also called spectrograms)  to be fed to the recurrent cell of the network. In Section \ref{ExpSec}, the actual STFT parameters utilized in our experiments, plus a visualization of the implemented frequency domain transform (see Figure \ref{stftFeat}) are provided.
 
 \subsection{Recurrent Cells}
 \label{rcell}
 A standard (or so-called ``vanilla'') RNN is basically a feed-forward neural network unrolled in time. It is fundamentally a set of weighted connections between a number of hidden states of the network and the same hidden states from the last time point, providing some sort of ``memory''. The challenge is that this memory is fundamentally limited in the same way that training very deep networks is difficult, due to factors such as vanishing gradients, hence limiting the memory of vanilla RNNs.
 
One solution to this elemental problem of RNNs, is creating so-called ``\emph{cell states}'' within the recurrent architecture that consist of a common thread through time, affected solely by linear operations at each time step. These recurrent cells can remember short-term memories for relatively longer sequences, mainly because the cell state connection to previous cell states is interrupted only by the linear elementary operations such as multiplication and addition.
 
 Two well-known examples of these recurrent architectures are Gated Recurrent Units (GRU) \cite{Cho_2014} and Long-Short Term Memory (LSTM) \cite{Hochreiter1997LongSM} cells. Although GRU is slightly simpler in its architecture, both cells are characterized by a persistent linear cell state surrounded by non-linear layers feeding input and parsing output from it. Technically, the cell state functions jointly with so-called ``gating'' layers that have the ability to ``forget'', ``update'' or ``reset'' the state of the cell, hence providing long or short-term memory capabilities. We benchmarked these two fundamental recurrent cells in our experiments. 
 
 GRU and LSTM, as a result of their architectural design, can only preserve information of the past (i.e., data from earlier in the input sequence), simply because they can only see input from the past. However, context in a sequence usually preserves information from both past and future. Although the linear dynamical system we use in this study is a causal system, an input/output pair of $\Delta$ seconds when observed anachronistically, may carry information in both directions. For example, we may observe and rationalize an output behavior only after we observe the continuity of the event in the output, which is an inconsequential correlation. For this reason, we also utilize the BiLSTM cell \cite{GRAVES2005602}, which provides, as the name implies, bidirectional input processing. Simply put, the input in a BiLSTM is fed twice, once from beginning to end and then in reverse from the end to the beginning, so the short- or long-term memory can be related both to and from the past.
 
All three of the aforementioned recurrent cells are benchmarked in our experiments in order to achieve a solid comparison of their capabilities in predicting dynamical system parameters.

\section{Experimental Setup}
\label{ExpSec}
In this section, we provide details of the experimental setup, starting with the dynamical system simulation module, which we used to create a dynamical input/output pairs dataset for different dynamical model parameters. Next, we provide the implementation details of how we constructed the frequency transform to be fed into the recurrent cells of the DRNN model. 

\subsection{Dynamical System Simulation Module}

In an actual physical scenario torque inputs can not be applied directly to a mechanical system such as the one described in the paper. Torque is applied via an actuator such as a motor. Accompanying the motor is also a motor driver card. There are transfer functions associated with both the motor and the driver. However, the motors are selected and the drivers are designed such that the bandwidths of the actuator and its driver are much higher than the mechanical system to be controlled. In addition, both of the transfer functions are usually a combination of first order filters with known (designed) parameters. 

Also, in an actual scenario, the angle $\theta$ is measured via sensors, such as encoders. Encoders can be digital or analog, but in either case, they are much faster (or selected to be faster) than the dynamical system on which they are mounted. Their transfer functions are usually provided by the manufacturer and for the encoder case, their bandwidths are much higher than the mechanical system.

Parameter identification of an actuator-mechanics-sensor system is usually performed by obtaining the frequency response. Sinusoidal inputs of various frequencies with known magnitudes are applied to the system, and steady state responses are logged. Based on the magnitude amplification and phase lag, transfer functions are identified. If designed carefully, this process provides the frequency response of the mechanical system and contains the data associated with the mechanical system. Actuator and sensor transfer functions are not captured in this process, largely because applying inputs at frequencies that excite these high frequency actuator and sensor dynamics is not possible or necessary. Hence, the experiments in the paper are constructed such that the low frequency, mechanical system dynamics are captured. Naturally, sensor noise impacts the data collected, and we have also included noise in our construction of the experiments.

A discrete (Tustin) standard second order system was used to generate the input-output data. The state-space description of the system is given as

\begin{align}
    x_{k+1}    = A x_k + B u_k \\
    \theta_{k} = C x_k + D u_k
\end{align}
where

\begin{align}
    A = \left[\begin{matrix}
        0 & 1\\[2mm]
     1-2\frac{\alpha^2+\wn^2}{M} &  2\frac{\alpha^2-\wn^2}{M}
    \end{matrix}\right],
    B = \left[\begin{matrix}
        0\\[2mm]
        1 
    \end{matrix}\right],    
\end{align}

\begin{align*}
    C = \frac{4\alpha\wn^2}{M^2}\left[\begin{matrix}
        \zeta\wn & (\alpha+\zeta\wn)
    \end{matrix}\right],
    D = \frac{\wn^2}{M},    
\end{align*}
$\alpha = 2/T_s$, and $M = \alpha^2+2\alpha\zeta\wn+\wn^2$.\\

Sampling time, $T_s = 0.001\;sec$, and natural frequency, $\omega_n = 1 \; rad/sec$ were kept fixed.

In order to construct the I/O sequence pair dataset to be used in our experiments, the system was initially excited with five different types of inputs, with eight different damping factor ($\zeta$) values applied for each input. Measurements were corrupted with normally distributed (Gaussian) noise.

The input signals (sampled at 1 KHz) were: a unit step input, a ramp input with a unit slope, and three sinusoids, which each had a magnitude of 10, and frequencies of 0.5, 1, and 2 Hz. Values for the damping factor, $\zeta$, ranged from \{$0.1, 0.2, \cdots, 0.8$\}. Ten seconds of data samples (i.e., 10sec$\times$1Khz = 10,001) were generated for each input/damping pair.

In total, using five different input types, eight distinct $\zeta$ values, and 7,001 overlapping 3-second sequences, we created a total of 280,040 (5$\times$8$\times$7,001) input/output sequence pairs and corresponding $\zeta$ values in our dataset.

\subsection{Frequency Domain Transform  of I/O Pairs}
As mentioned in the previous subsections, we fed complex spectrograms of the collected I/O time sequence pairs into our DRNN. In this subsection we provide details of the implementation of these spectrogram-based features. The I/O pairs we utilized were of $\Delta\cdot f_{s}$ = 3,000 size, as $\Delta$ was selected as 3 seconds and $f_{s}$ as 1 kHz in our experiments. The size of the window function of the STFT was selected as $L = k\cdot\Delta\cdot f_{s}$ = 2 seconds length (i.e., k = 2/3). The overlap ratio $p$ was selected as 0.95, therefore creating 11 long spectrograms in the time dimension.

\begin{figure*}[t]
\centering
\begin{subfigure}{2.56cm}	
	\includegraphics*[trim= 0 0 0 0,clip=true,width=1\textwidth]{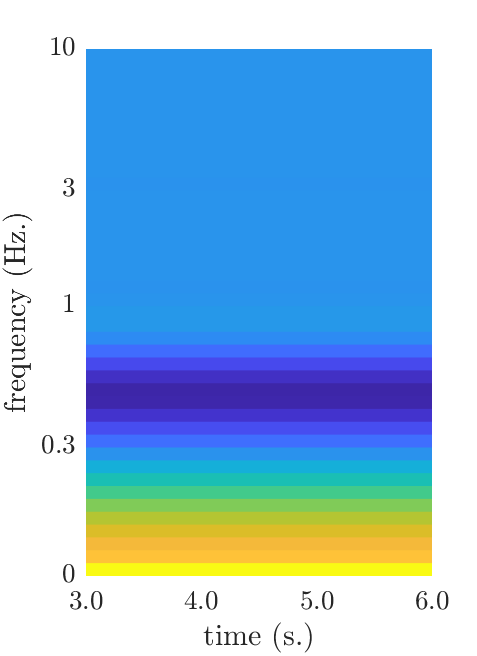}
	\caption{\emph{real}(Input)}	
\end{subfigure}
\begin{subfigure}{2.94cm}
\centering
	\includegraphics*[trim= 0 0 0 0,clip=true,width=1\textwidth]{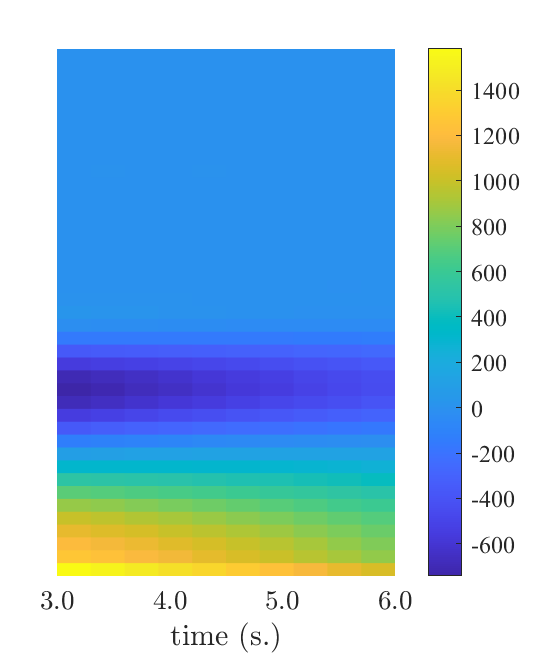}
	\caption{\emph{real}(Output)}		 
\end{subfigure}
\begin{subfigure}{2.56cm}	
	\includegraphics*[trim= 0 0 0 0,clip=true,width=1\textwidth]{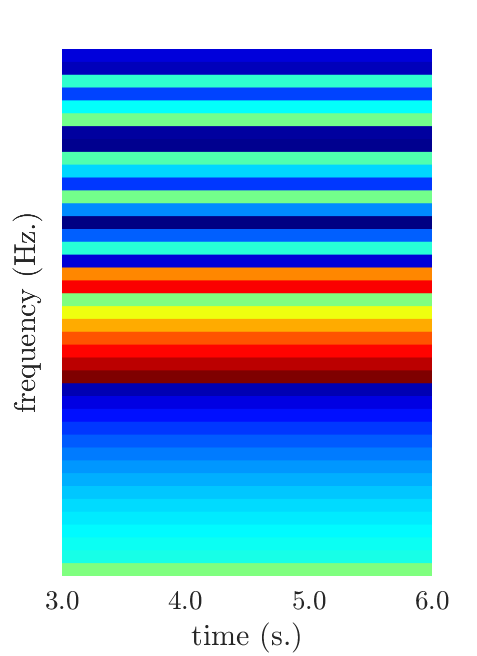}
	\caption{\emph{phase}(Input)}		
\end{subfigure}
\begin{subfigure}{2.94cm}
\centering
	\includegraphics*[trim= 0 0 0 0,clip=true,width=1\textwidth]{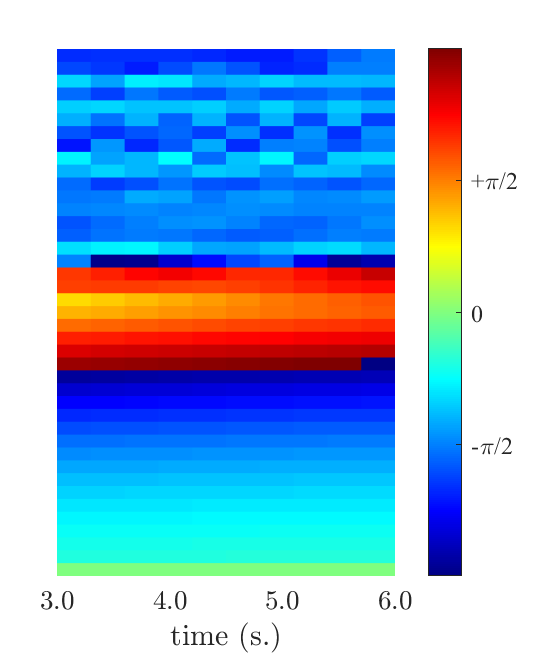}
	\caption{\emph{phase}(Output)}		 
\end{subfigure}
\caption{The figures each depict a visual representation of the feature vector fed into the recurrent cells.  
}
 \label{stftFeat}
\end{figure*}

The frequency range of the spectrograms are limited to between 0 and 10 Hz, and the selection of the frequency range is down to the frequency content of the system and the input signals. The natural frequency of the system considered is approximately 0.159 Hz, and the highest frequency used in the input is 2 Hz. The range of $[0, 10]$ Hz is selected in order to include the frequency content of the system in consideration. It should be noted here that the frequency range of the spectrograms would be different if another system or another input sequence was used. The intermediate frequencies in this range were sampled exponentially, so that the frequency dimension of the spectrograms become a log-frequency axis. The sampling is carried out using the formula:

\begin{equation}\label{stftfreq}
    f[h]=\sum_{h=0}^{N-2}{10^{\frac{h}{20}-1}}
\end{equation}   

Using the 41 frequencies calculated using Equation \ref{stftfreq} and appending 0Hz, we obtained a N = 42 long frequency dimension in the spectrograms\footnote{Such that the selected frequencies are [0Hz, 0.1Hz, 0.1122Hz, 0.1259Hz, ... 7.9433Hz, 8.9125Hz, 10Hz]}. Using the two complex $42\times11$ spectrograms, we then created a feature vector. Since the RNN model we propose can only have real weights and activations, we extracted a real valued feature vector from the spectrograms using the formula:

\begin{equation}\label{featvec}
    \begin{aligned}
    &    v = [ && real(input_{STFT}], real(output_{STFT}) ... \\
    &          && phase(intput_{STFT}), phase(output_{STFT})]
    \end{aligned}
\end{equation}

In Equation \ref{featvec}, $real(\cdot$) denotes an operator that extracts the real part of a complex number and $phase(\cdot$) denotes an operator that extracts the phase angle of a complex number in radians. Consequently, the feature vector $v$ becomes (42$\times$4)$\times$11 tensor, as shown in Table \ref{DeepLModel}.

Figure \ref{stftFeat} depicts a visual representation of the proposed feature vector. The complex spectrograms obtained from 3-seconds-long input/output sequences are of 42$\times$11 size.  The real parts of the input and output spectrograms are shown in (a) and (b); whereas (c) and (d) depict the phase angles of the input and output spectrograms, respectively using false color. The specific example shown in Figure~\ref{stftFeat}, is 3 to 6 seconds of a ramp input fed into our system with a damping coefficient ($\zeta$) of 0.1.

\subsection{Experiments}
Two sets of cross-validation experiments were conducted. The first set of experiments were two-fold cross validation experiments, in which the entire 280,040 I/O sequence pairs were divided into two sets according to their $\zeta$ values. For each fold, one half was used for testing, whilst the other half as used for training and validation. For the first fold, the $\zeta$ values in the training + validation sets were $\zeta = \{0.1, 0.3, 0.5, 0.7\}$, whereas the test set included sequences with $\zeta$ values $\{0.2, 0.4, 0.6, 0.8\}$. For the second fold, as expected, the $\zeta$ values were interchanged between the test and training sets. In practice, the significance of this experiment was assessing the ability of the system to predict the $\zeta$ value, which was not used during training. In other words, we aimed at testing the ability of the model to predict an unseen $\zeta$ value. This set of experiments are referred to as ``½ - sep. $\zeta$'', denoting the two-fold experiment with ``\emph{separate}'' $\zeta$ values in the test and training sets.

The second set of experiments were also two-fold cross validation experiments, in which the entire 280,040 I/O sequence pairs were divided into two random non-intersecting sets. Again, for each fold, one half was used for testing and the other half for training and validation. In this set of experiments, the dataset was divided randomly, hence each set consisted of an equal distribution of $\zeta$ values, and the models were trained and could be tested with any input type and $\zeta$ values in the dataset. However, as expected, the testing and training sets did not include identical sequences (although they included partially overlapping sequences). Our intention in these experiments was to assess the ability of the system to predict a previously seen $\zeta$ value from a previously (partially) unseen I/O sequence pair. These set of experiments is referred to as ``½ - mix. $\zeta$'', denoting the two-fold experiment with ``\emph{mixed}'' $\zeta$ values in the testing and training sets.

Both sets of experiments were then rerun using a different recurrent cell each time. Having three fundamental recurrent cells (i.e., GRU, LSTM, and BiLSTM) as our benchmark, we initially conducted a total of six experiments, namely Exp.1, Exp.2 and Exp.3 belonging to the first set of experiments; whereas Exp.4, Exp.5 and Exp.6 belonging to the second set of experiments. An additional variant of Exp.6, namely Exp.6b (which also utilizes BiLSTMs) was also conducted for further analysis. Same hyper-parameters were used in all experiments. Stochastic gradient descent with momentum (0.9) was used as the optimizer for all. The initial learning rate was set to 5$\cdot10^{-4}$. All the aforementioned experiments were trained for 45 epochs. At each 15 epoch interval, the learning rate was dropped by a factor of 0.1.  

\begin{table}[t]
\centering
\begin{tabular}{|l|l|l|c|c|}
\hline
\multicolumn{1}{|l|}{Exp. No.} & Set& \multicolumn{1}{l|}{RNN Cell} & \begin{tabular}[l]{@{}c@{}}train.\\ mad.\end{tabular} & \begin{tabular}[l]{@{}l@{}}test.\\ mad.\end{tabular} \\ \hline
Exp.1& ½ - sep. $\zeta$ & GRU & 0.0200 & 0.0755\\ \hline
Exp.2& ½ - sep. $\zeta$ & LSTM& 0.0240 & 0.0790\\ \hline
Exp.3& ½ - sep. $\zeta$ & BiLSTM  & 0.0145 & 0.0645\\ \hline
Exp.4& ½ - mix. $\zeta$& GRU & 0.0310 & 0.0325\\ \hline
Exp.5& ½ - mix. $\zeta$ & LSTM& 0.0310 & 0.0325\\ \hline
Exp.6a& ½ - mix. $\zeta$ & BiLSTM  & 0.0200 & 0.0210\\ \hline
Exp.6b& ½ - mix. $\zeta$ (Step Input @ 3-6sec.) & BiLSTM  & - & 0.0097\\ \hline \hline
Exp.7& ½ - mix. $\zeta$ (150 ep. - Ext. Data.) & BiLSTM  & 0.014 & 0.015\\ \hline
\end{tabular}
\caption{Mean absolute deviation (mad) of $\zeta$ prediction errors for all experiments}
\label{ResultsTable}
\end{table}

\begin{figure}[t]
\centering
\begin{subfigure}{3.90cm}	
	\includegraphics*[trim= 0 0 0 0 ,clip=true,width=1\textwidth]{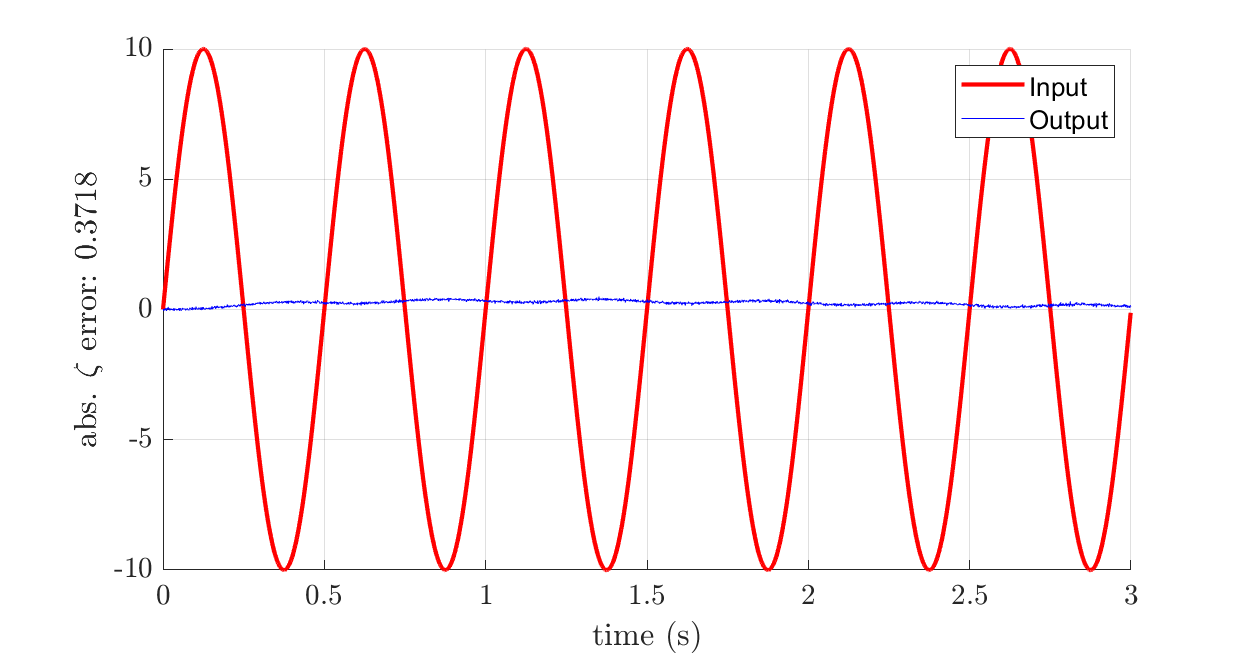}
	\caption{}
\end{subfigure}
\begin{subfigure}{3.90cm}	
	\includegraphics*[trim= 0 0 0 0 ,clip=true,width=1\textwidth]{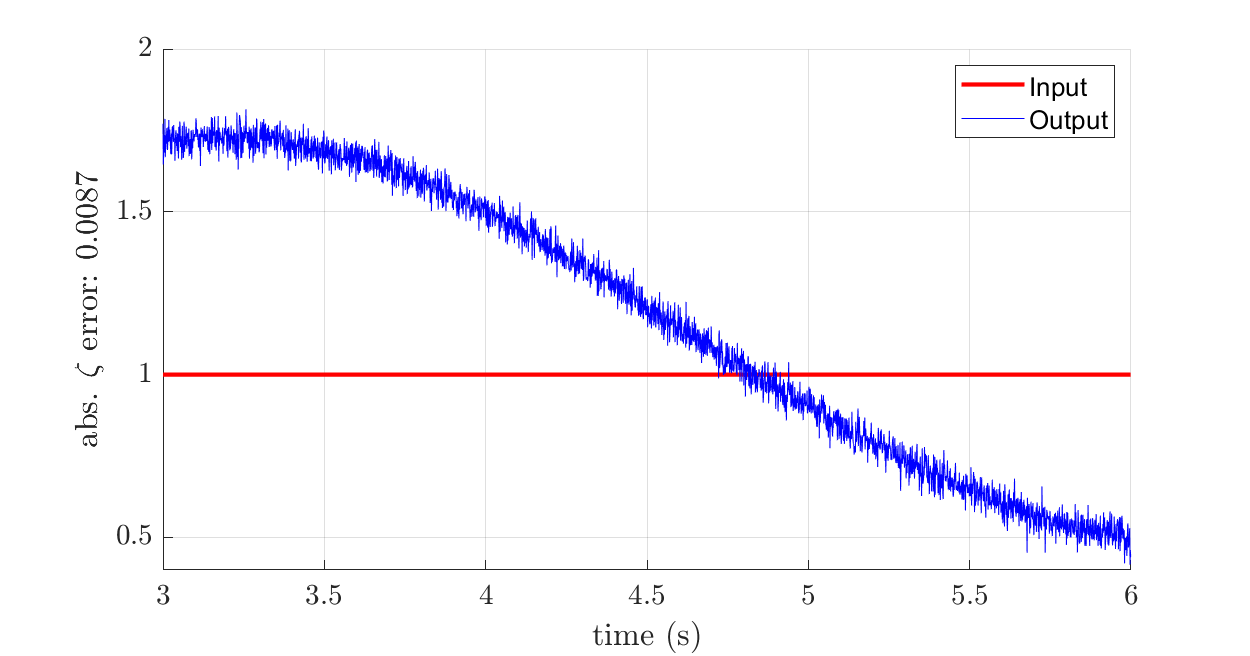}
	\caption{}
\end{subfigure}
\begin{subfigure}{3.90 cm}	
	\includegraphics*[trim= 0 0 0 0 ,clip=true,width=1\textwidth]{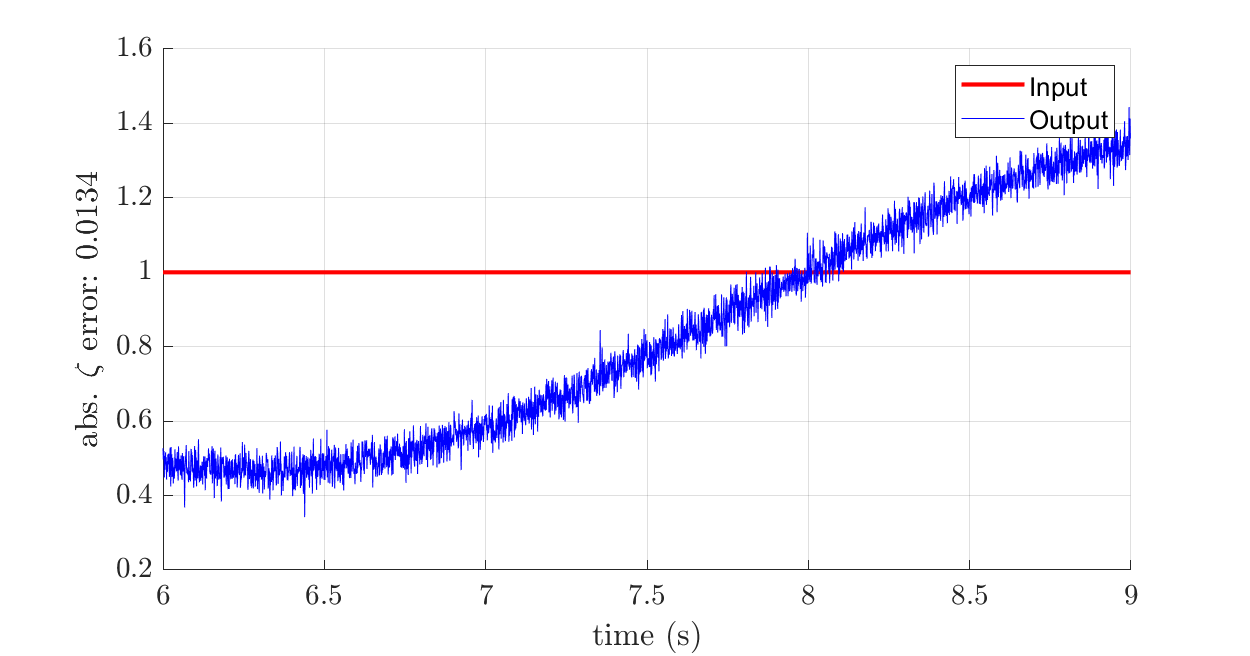}
	\caption{}
\end{subfigure}
\caption{Figures show sample I/O sequences, with $\zeta$ prediction errors provided in captions for each figure. 
}
 \label{SampleResults}
\end{figure}

\section{Results}
We commence with the results of the ``½ - sep. $\zeta$'' experiments, which are presented as Exp.1, Exp.2, and Exp.3 in Table \ref{ResultsTable}. In all experiments, the mean-absolute-deviation (MAD) of the $\zeta$ errors obtained from the separate folds are averaged. For the ``½ - sep. $\zeta$'' experiments, the model with BiLSTM recurrent cell performed the best, with a MAD-$\zeta$ error of 0.0645 for the test set, whereas the results for the models with GRU and LSTM cells exhibited a relatively poorer level of performance compared to models with BiLSTM. A MAD-$\zeta$ error of 0.0645 is, in practical terms, acceptable considering the tested $\zeta$ values do not exist in the training set and their range is between 0.1 and 0.8. Table \ref{ResultsTable} also presents the MAD $\zeta$ error for the training set to aid evaluation of the overfitting extent that occurred during training.

The `½ - mix. $\zeta$'' experiments, which are presented as Exp.4, Exp.5, and Exp.6a in Table \ref{ResultsTable} exhibited better performance, when compared to Exp.1, Exp.2, and Exp.3. Among these experiments, the model with BiLSTM recurrent cell, trained in Exp.6a performed the best, with a MAD-$\zeta$ error of 0.021 for the test set, whereas the results for the models with GRU and LSTM cells performed relatively poorer. From this result, we can draw two main conclusions. First, if we train the system with all types of $\zeta$, the model is capable of predicting the dynamical system parameter with a very high degree of accuracy. Second, BiLSTM always performs better when compared to both GRU and LSTM, denoting that context within a dynamical system sequence model should be classified in a bidirectional manner in time. We consider this to be a significant outcome of the experiments, considering that, to the best of our knowledge, there is no empirically stated emphasis on the benefits of utilizing BiLSTMs for dynamical systems identification problem in the literature.

The experiments in Table \ref{ResultsTable} were focused on the effect of using different $\zeta$ values during the training. In order to answer questions such as ``\emph{What type of input should be fed into the system to better predict a dynamic parameter?}'' or ``\emph{Which time interval of the I/O sequence is more effective in predicting the dynamic system parameter?}'', we conducted some additional experiments. In order to understand the performance provided by the experiments, we first analysed sample cases presented in Figure \ref{SampleResults}. For example in Figure \ref{SampleResults}a, a 2 Hz sinusoidal input to a system with $\zeta$ = 0.8 and an output response for 0-3 seconds can be seen. After feeding the depicted I/O sequence pair into the model obtained from Exp.6 (which performed the best among Exp.1-6), a $\zeta$ value of 0.4282 was predicted. when compared to the actual value of $\zeta$ = 0.8, this represented a very poor level of performance. 

\begin{figure*}[t]
\centering
\begin{subfigure}{3.9cm}	
	\includegraphics*[trim= 180 0 145 40,clip=true,width=1\textwidth]{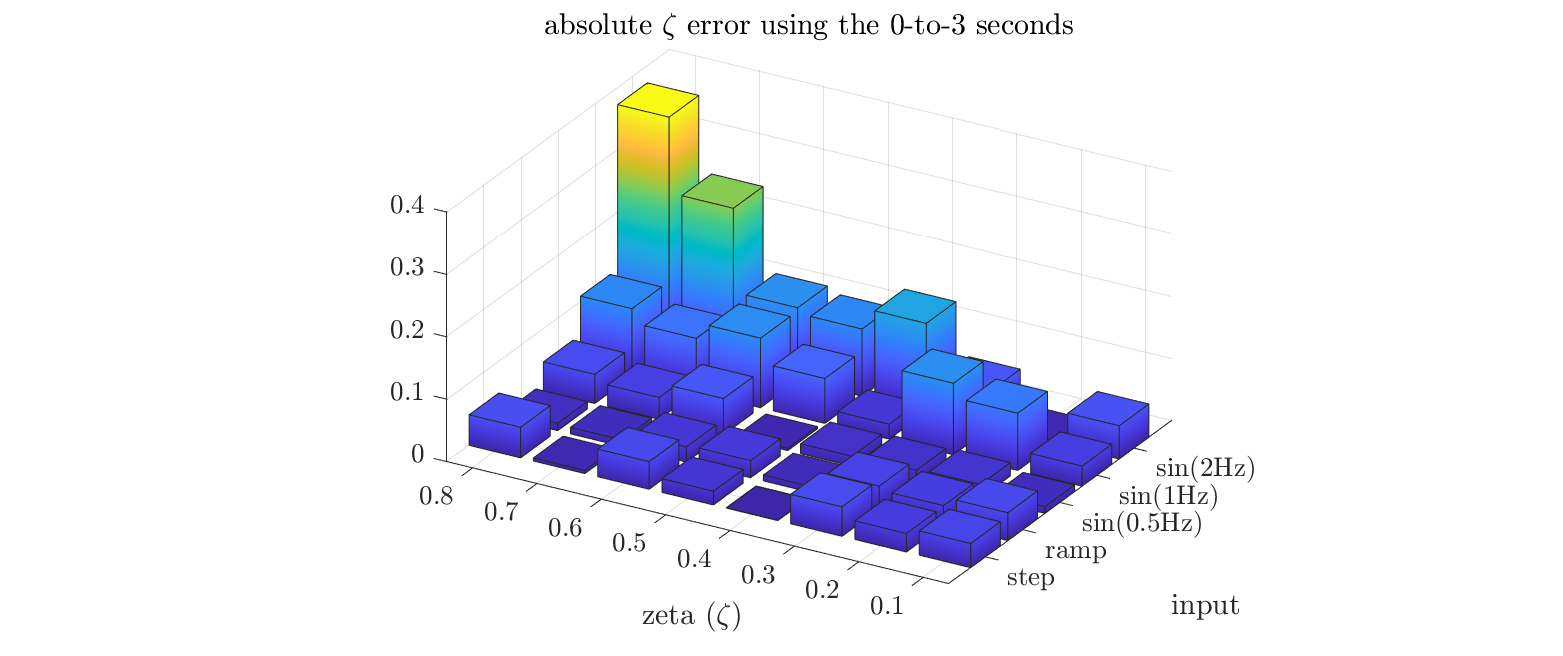}
	\caption{I/O for 0-3 sec.}	
\end{subfigure}
\begin{subfigure}{3.9cm}	
	\includegraphics*[trim= 180 0 145 40,clip=true,width=1\textwidth]{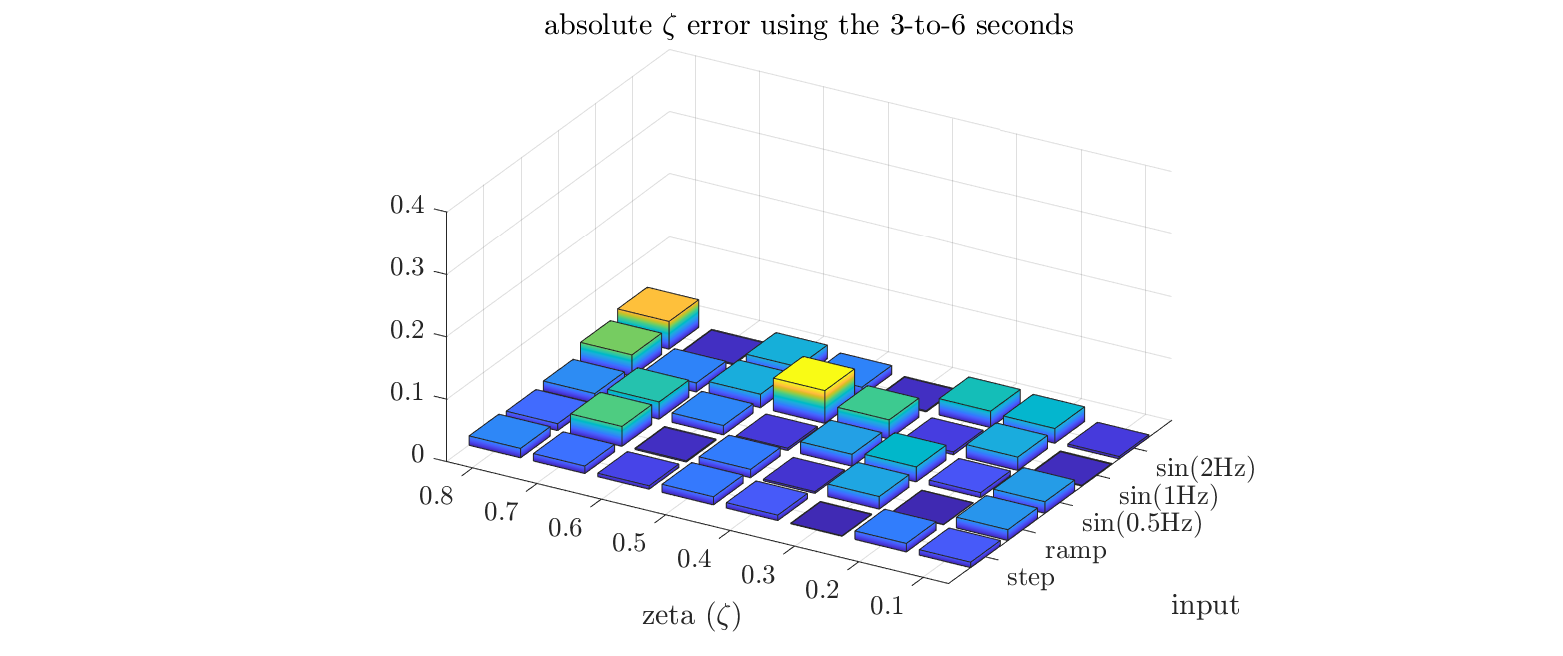}
	\caption{I/O for 3-6 sec.}	 
\end{subfigure}
\begin{subfigure}{3.9cm}	
	\includegraphics*[trim= 180 0 145 40,clip=true,width=1\textwidth]{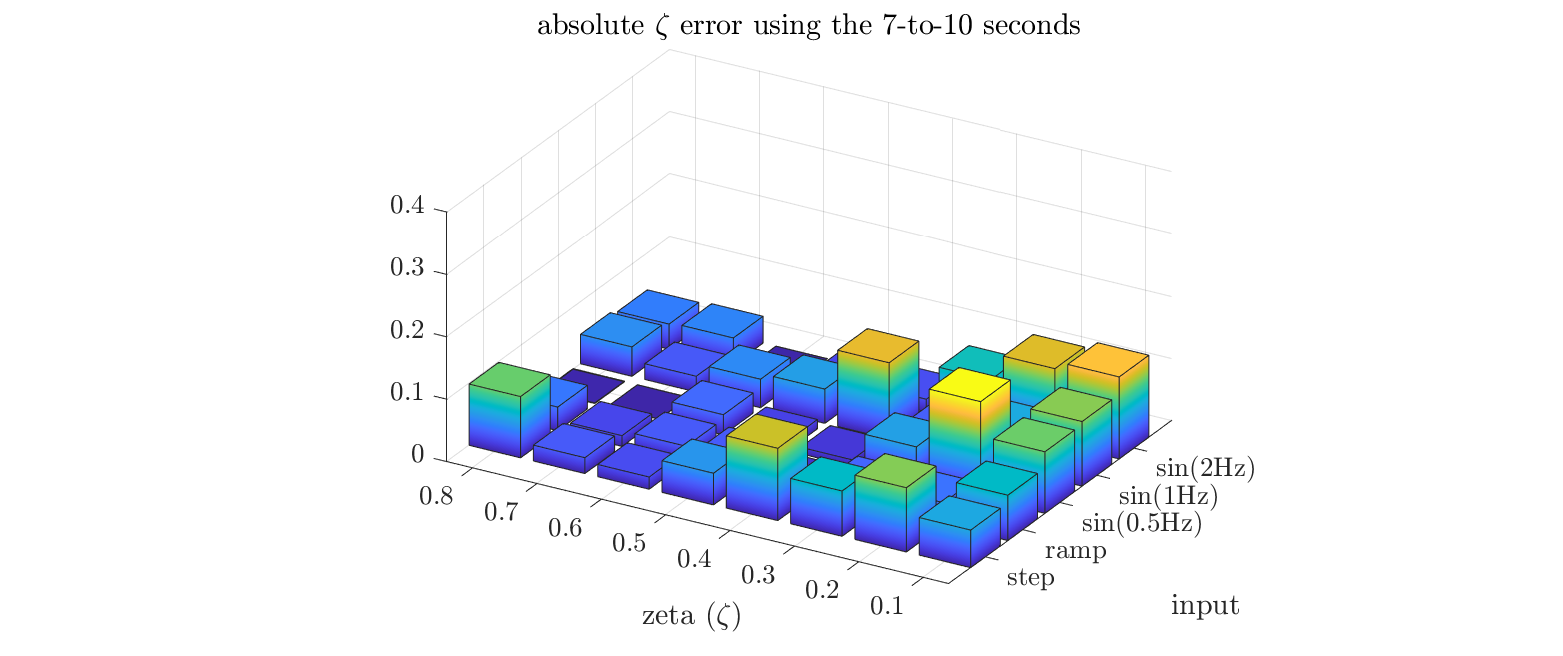}
	\caption{I/O for 7-10 sec.}	 
\end{subfigure}
\caption{The figures show 2D histograms of how much ``\textbf{absolute $\zeta$ prediction error}'' was obtained for different input types and different damping levels, using the \textbf{Exp.6} model.
}
 \label{AbsZeta2dHist}
\end{figure*}

However when we observe Figure \ref{SampleResults}b (or similarly Figure \ref{SampleResults}c), the performance seen was significantly better. For a step input to a system with $\zeta$ = 0.1 and using the output response of 3-6 seconds, the predicted $\zeta$ value (again using the model trained in Exp.6) was 0.1087 (i.e., with an absolute error of 0.087). This achieved a considerably better result.

The varying performance of the sample cases demonstrates that the prediction performance relies upon the type of input used, or the time interval of the response of the system that is fed into the deep learning model. In order to analyze this variance, Figure \ref{AbsZeta2dHist} shows 2D histograms of how much ``\textbf{absolute $\zeta$ prediction error}'' was obtained for different input types and different damping levels, using the model obtained in Exp.6. The x-axis in these histograms presents the different input types, whilst the y-axis presents the different $\zeta$ values used in the testing. For example, Figure \ref{AbsZeta2dHist}a depicts the 2D histograms of absolute $\zeta$ error when 0-3 seconds of I/O sequence pairs were fed into the DRNN model; whereas, Figures \ref{AbsZeta2dHist}b and \ref{AbsZeta2dHist}c depicts the corresponding 2D histograms when 3-6 seconds and 7-10 seconds of the I/O sequence pairs were fed in, respectively. For example, it can be seen from Figure \ref{AbsZeta2dHist}a that using a 2 Hz input, and trying to predict the parameter of a dynamic system with $\zeta$ = 0.8 (i.e., highly dampened) in the 0-3 second interval was prone to a high degree of errors. We actually observed this in Figure \ref{SampleResults}a, hence Figure \ref{AbsZeta2dHist}b shows that the best results (i.e., the lowest MAD-$\zeta$ errors) were obtained when 3-6 seconds of I/O sequence pairs were fed into the system. For this interval only, and for the step-input only, we calculated the MAD-$\zeta$ error of the entire test set for all $\zeta$ values using the Exp.6 model, and presented this result in another row of Table \ref{ResultsTable} as Exp.6b. The MAD-$\zeta$ error for this interval was 0.0097, which was the best-case scenario in our ``45-epoch'' experiments.

It should also be noted here that the effect of a small $\zeta$ (lightly damped) was expected to be more easily distinguishable, since it would exhibit a larger magnitude in the oscillation. As the $\zeta$ was increased, the time responses should tend to be more similar to each other, and hence would make it harder for the deep learning algorithm to converge to the true value of $\zeta$. This expectation was most notable as can be seen in Figure~\ref{SampleResults}a, specifically when the input signal was a sinusoid at 2 Hz. Since, the system is a second-order, low-pass filter, a sinusoid at 2 Hz will be attenuated more than a sinusoid at a lower frequency, i.e., the magnitude of the output signal would be smaller, and hence the signal-to-noise ratio would be smaller. A smaller signal-to-noise ratio was expected to cause problems for the deep learning algorithm.

\subsection{Additional Experiments using Step Inputs with Varying Magnitudes}
After analyzing the results of the first two sets of experiments,and witnessing the relatively superior performance of BiLSTM-based models, we have carried out another experiment, with an extended dataset. The new dataset included additional step input signals of varying magnitudes, again sampled at 1 KHz. In addition to our original input set (i.e. a unit step input, a ramp input with a unit slope, and three sinusoids, which each had a magnitude of 10, and frequencies of 0.5, 1, and 2 Hz), we have added three step input sequences with magnitudes -1, +10 and -10. For each input we again ha damping factor, $\zeta$, vaues ranged from \{$0.1, 0.2, \cdots, 0.8$\}. Similarly, ten seconds of data samples (i.e., 10sec$\times$1Khz = 10,001) were generated for each input/damping pair. Hence in total, using eight different input types, eight distinct $\zeta$ values, and 7,001 overlapping 3-second sequences, the extended dataset included a total of 448,064 (8$\times$8$\times$7,001) input/output sequence pairs and corresponding $\zeta$ values.

\begin{figure*}[t]
\centering
\begin{subfigure}{3.8cm}	
	\includegraphics*[trim= 180 0 145 40,clip=true,width=1\textwidth]{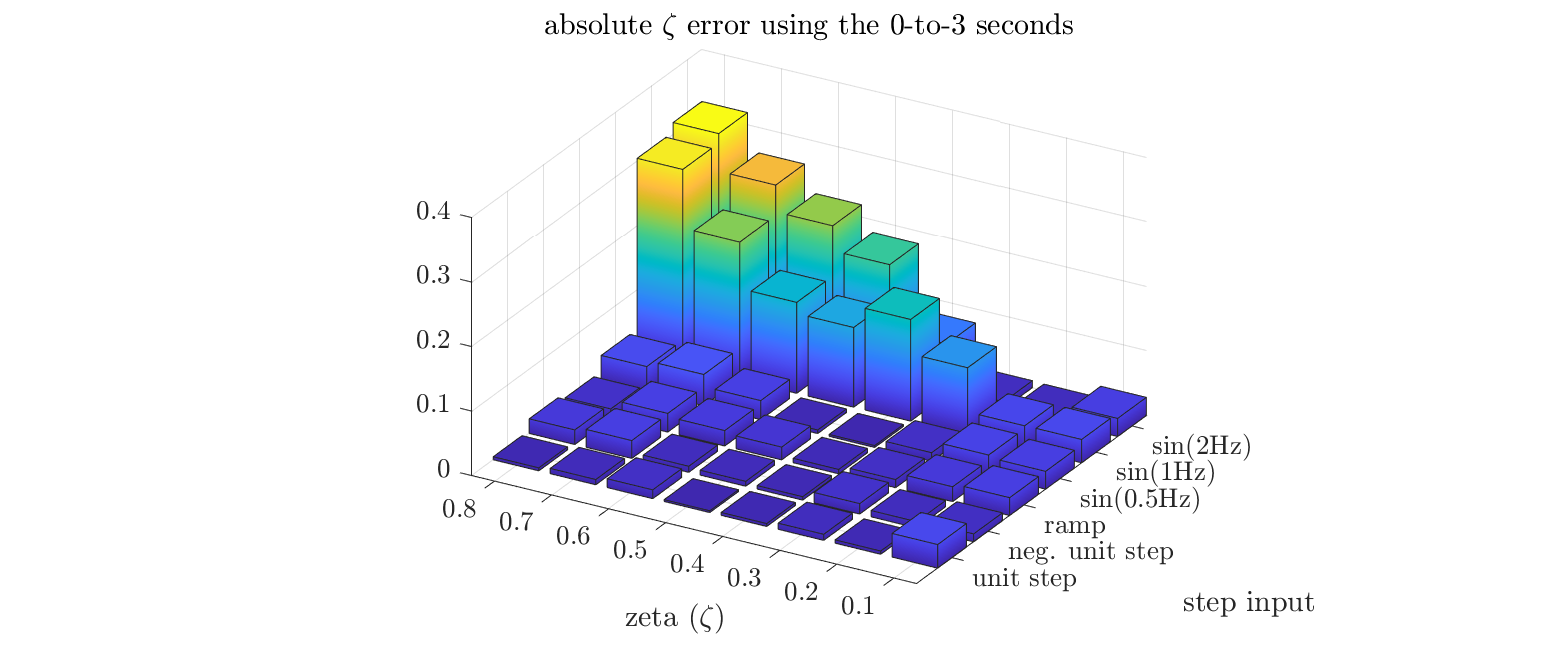}
	\caption{I/O for 0-3 sec.}	
\end{subfigure}
\begin{subfigure}{3.8cm}	
	\includegraphics*[trim= 180 0 145 40,clip=true,width=1\textwidth]{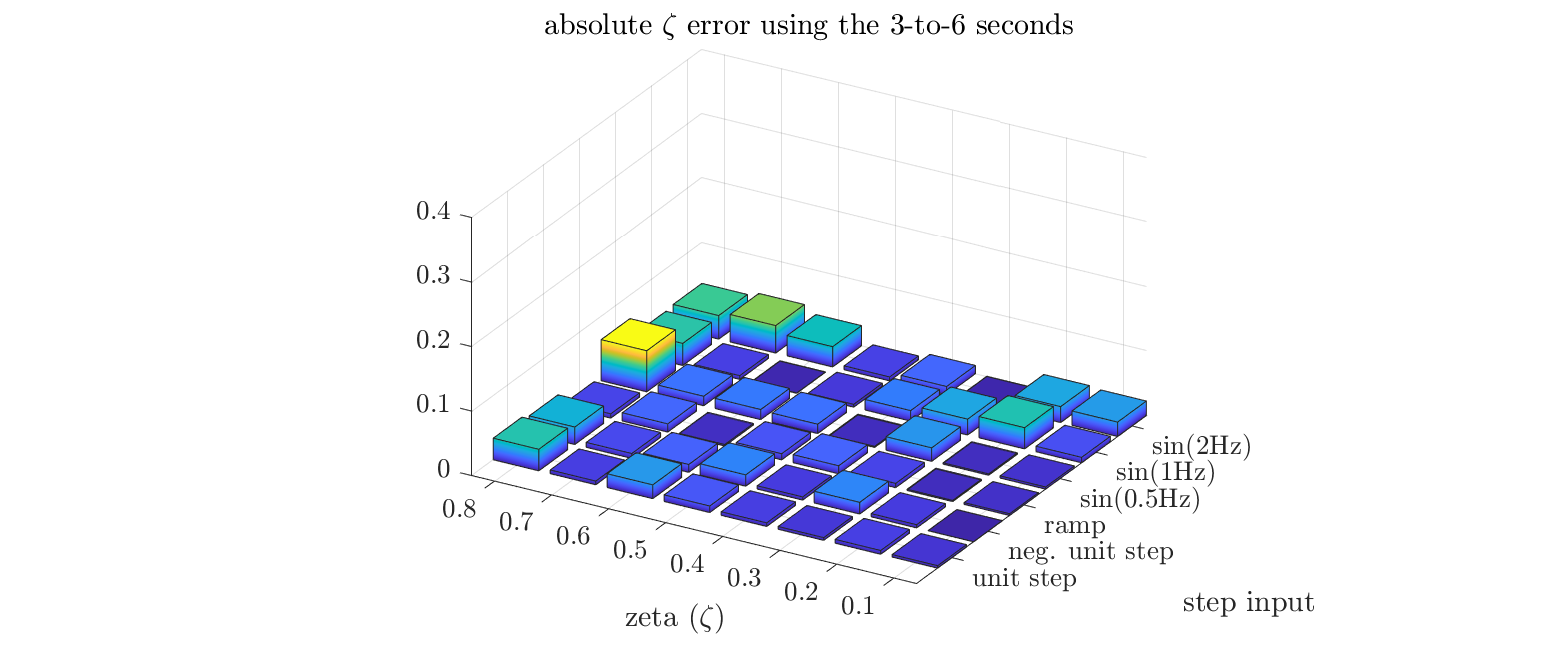}
	\caption{I/O for 3-6 sec.}	 
\end{subfigure}
\begin{subfigure}{3.8cm}	
	\includegraphics*[trim= 180 0 145 40,clip=true,width=1\textwidth]{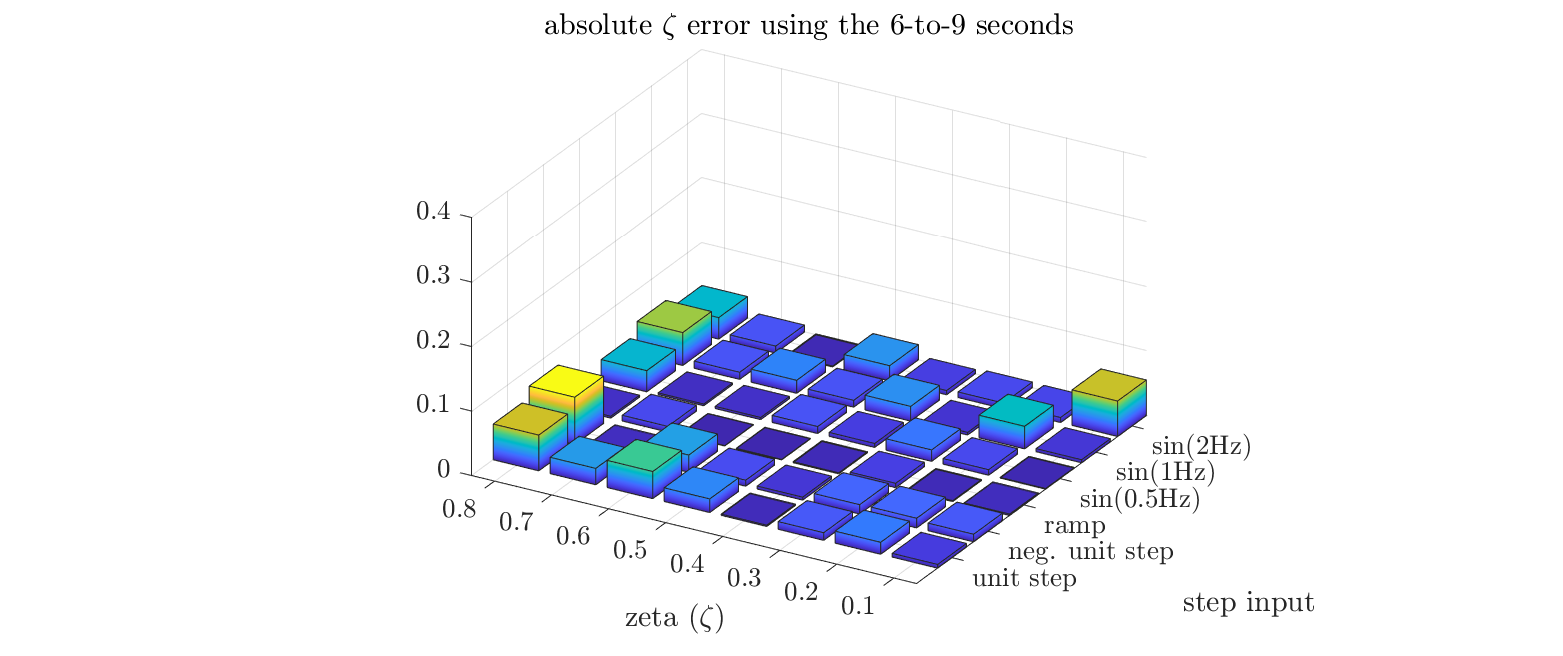}
	\caption{I/O for 6-9 sec.}	 
\end{subfigure}
\caption{The figures show 2D histograms of how much ``\textbf{absolute $\zeta$ prediction error}'' was obtained for different input types and different damping levels, using the \textbf{Exp.7} model.
}   
 \label{AbsZeta2dHist2}
\end{figure*}

Similarly to Exp.4-6, the extended dataset were divided into two random non-intersecting sets for a two-fold cross-validation. The same hyper-parameters were applied except for the fact that this new network was trained for 150 epochs, unlike the previous experiments that were all trained for 45 epochs. After the 45\textsuperscript{th} epoch, the learning rate drop factor was manipulated manually for best performance. This experiment is referred to as Exp.7 and can be seen in the final row of Table \ref{ResultsTable}.

In Figure \ref{AbsZeta2dHist2}, we present absolute $\zeta$ prediction error histograms for Exp.7 model, similarly to Figure \ref{AbsZeta2dHist}, which was prepared using Exp.6 model. When compared to Figure \ref{AbsZeta2dHist}, the error rates in Figure \ref{AbsZeta2dHist2} are dramatically better for all input types, including the additional input signals introduced in the extended dataset. The prediction errors are almost always under 0.1, except for the case, where sinusoidal inputs and their initial response (i.e. between 0-3 sec.) are applied to a high friction model. As explained above, this is an acceptable result considering that high friction systems will inevitably lag the response of a high frequency input. 

\begin{figure*}[t]
\centering
\begin{subfigure}{3.8cm}	
	\includegraphics*[trim= 210 0 195 40,clip=true,width=1\textwidth]{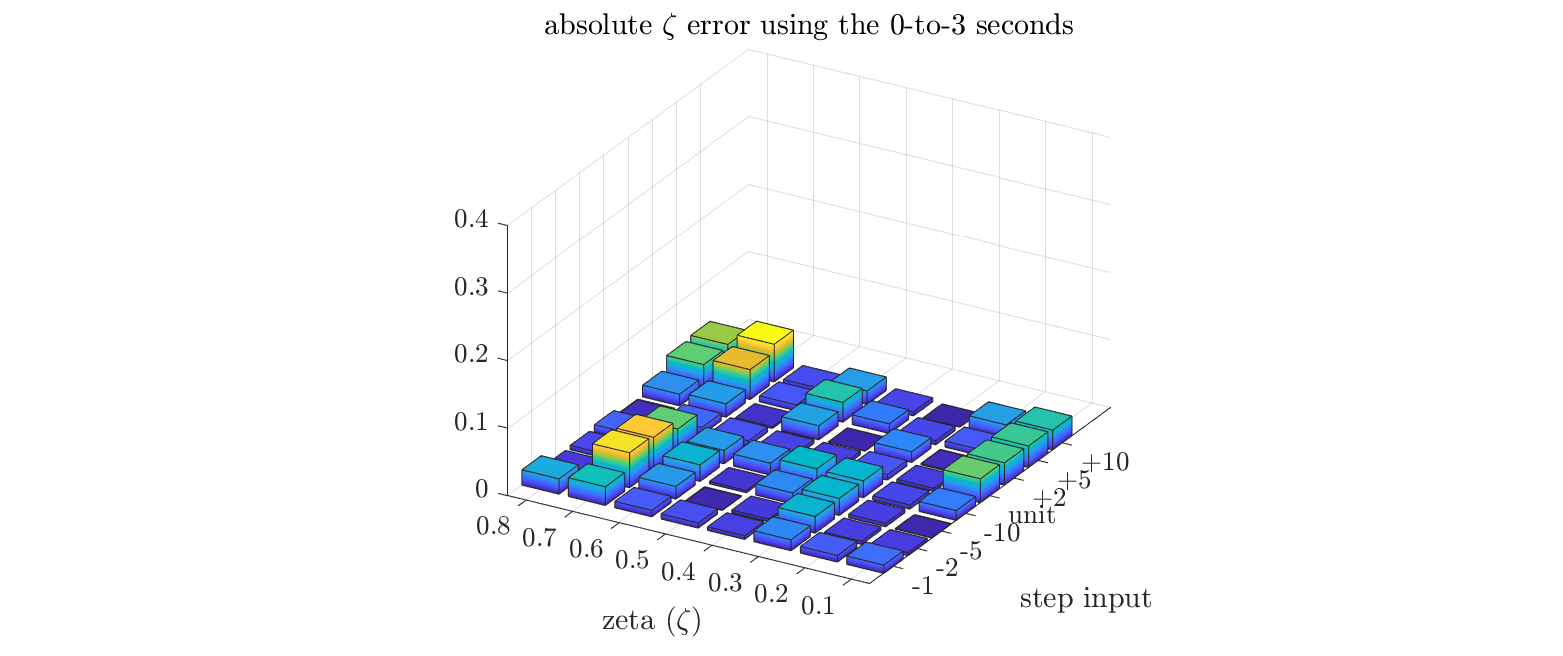}
	\caption{I/O for 0-3 sec.}	
\end{subfigure}
\begin{subfigure}{3.8cm}	
	\includegraphics*[trim= 210 0 195 40,clip=true,width=1\textwidth]{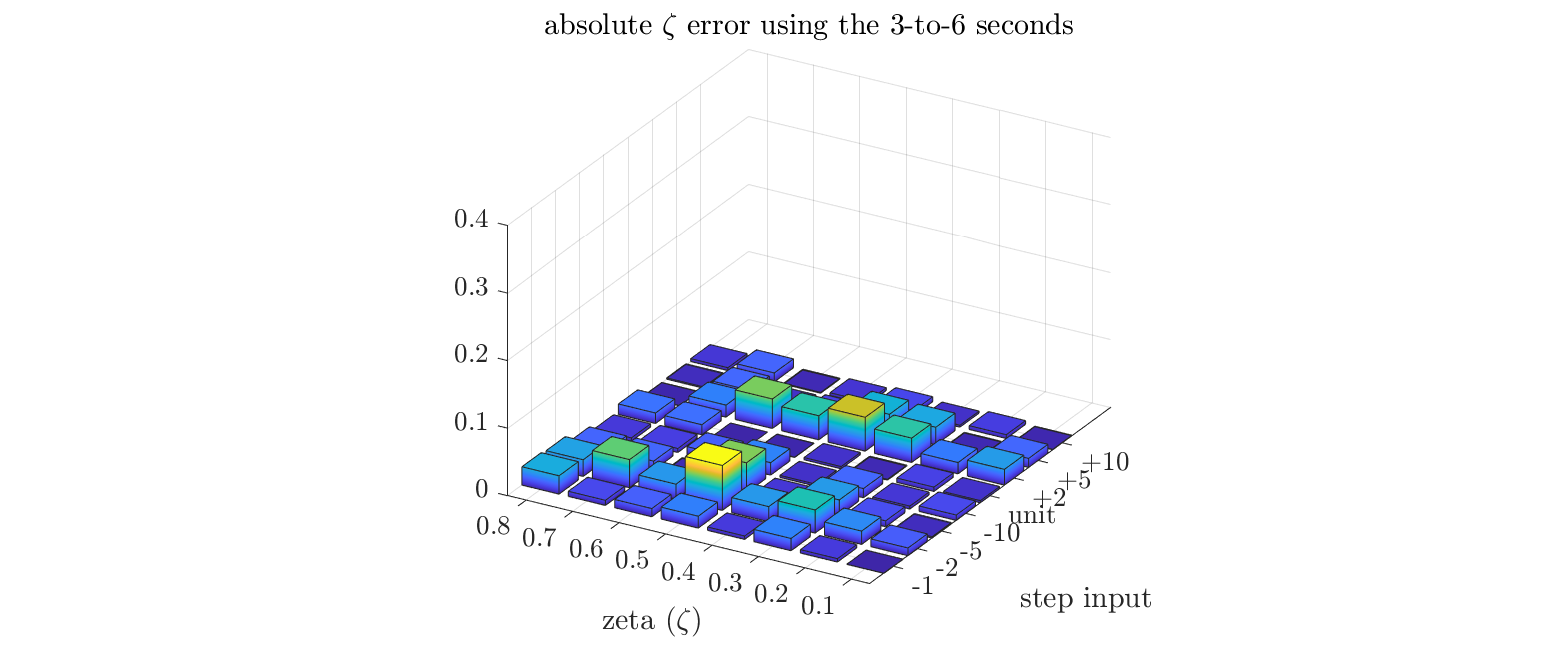}
	\caption{I/O for 3-6 sec.}	 
\end{subfigure}
\begin{subfigure}{3.8cm}	
	\includegraphics*[trim= 210 0 195 40,clip=true,width=1\textwidth]{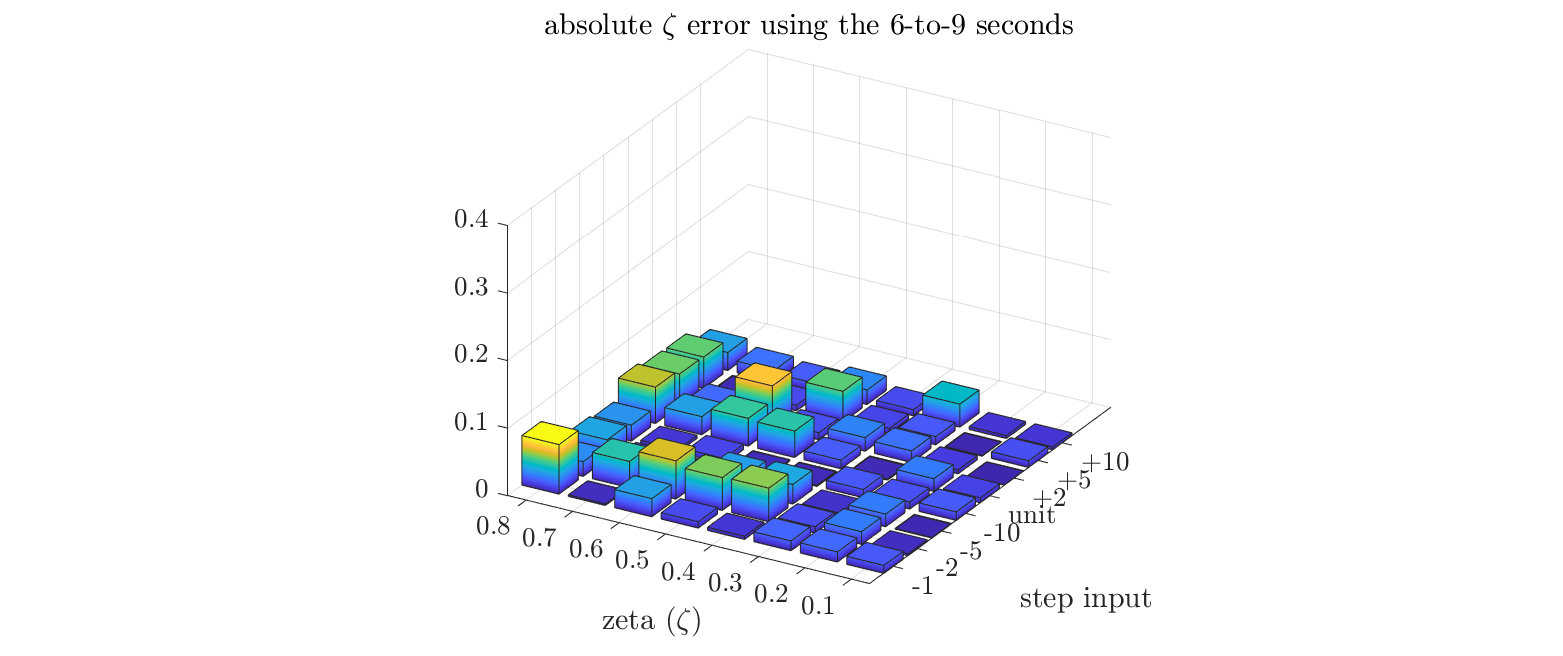}
	\caption{I/O for 6-9 sec.}	 
\end{subfigure}
\caption{The figures show 2D histograms of how much ``\textbf{absolute $\zeta$ prediction error}'' was obtained for different \textbf{step input} types and different damping levels, using the \textbf{Exp.7} model.
}   
 \label{AbsZeta2dHist3}
\end{figure*}

Furthermore, we analysed the prediction capabilities of the model trained in Exp.7, against step inputs of varying magnitudes, including values which were not introduced during training. In Figure \ref{AbsZeta2dHist3}, absolute $\zeta$ prediction error histograms for step inputs of varying magnitudes can be seen. In this figure, in addition to step inputs with unit, negative unit, -10 and +10 magnitudes that were used in training Exp.7 model; -2,+2,-5,+5 magnitude step inputs are also tested. It is celarly seen in Figure \ref{AbsZeta2dHist3} that, although the second group of step inputs were not introduced during training, the model trained in Exp.7 can successfully generalize different magnitude step input sequences and predict the related dynamical systems parameter with very high accuracy.

\section{Conclusions and Future Work}
The main goal of this paper was to find an effective deep recurrent neural architecture for the al systems parameter identification task. For this purpose, three gated recurrent cells, namely GRU, LSTM, and BiLSTM were comparatively experimented. The results showed that at late transient, by feeding a step input into the system and by utilizing BiLSTM recurrent cells in the proposed 6-layered neural architecture, we achieved the lowest error rates in predicting the damping coefficient of a second order linear dynamic system. The average test MAD for our best experiment (Exp.7), which was trained longer with an extended dataset, was significantly better when compared to previous experiments; showing us that the models are open to further improvement in performance and more importantly as we add new types of inputs, the machine learning model has the capacity to generalize them without any loss of accuracy. 

We principally observe two significant outcomes in our experiments. First, BiLSTM cells with bidirectional gradient flow performed better than single-direction gated recurrent cells (GRU and LSTM in our case); thus showing that context within a dynamical system sequence model correlated in a bidirectional manner, when it comes to the systems identification problem. Second, through experimentation, we investigated at which exact instant and with what kind of an input the system should be excited  in order to obtain the best parameter identification results.

Future research directions point to a strong collaboration between the fields of automatic control systems engineering and deep learning. We believe that the more these fields work in partnership, the greater the potential impact that they will have in transforming each other’s research directions. A good next step, therefore, would be to embed a deep learning-based parameter identification system, such as the one proposed in the current study, into the actual closed loop of the dynamic system, thus providing parameter-aware automated control.


\bibliographystyle{plain}      
\bibliography{ref}

\end{document}